%% file: main.tex
\pdfoutput=1

\documentclass[11pt]{article}

\usepackage[]{acl}

\usepackage{times}
\usepackage{latexsym}

\usepackage[T1]{fontenc}

\usepackage[utf8]{inputenc}

\usepackage{microtype}

\usepackage{inconsolata}

\usepackage[ruled, vlined, linesnumbered]{algorithm2e}
\usepackage{bm}
\usepackage{bbm}
\usepackage{amsthm}
\usepackage{amsmath}
\usepackage{mathrsfs} 
\usepackage{tabularx}
\usepackage{graphicx}
\usepackage{multirow}
\usepackage{xspace}
\usepackage{xcolor}
\usepackage{enumitem}
\usepackage{placeins}
\usepackage{soul}
\usepackage{mathtools}
\usepackage{booktabs}
\usepackage{hyperref}
\usepackage{subcaption}
\usepackage{pgfplots}
\usepackage{csquotes}
\usepackage{cleveref}
\usepackage{balance}
\usepackage{multicol}
\usepackage{makecell}
\usepackage{amssymb}
\usepackage[switch]{lineno}
\DeclareUnicodeCharacter{2212}{−}
\usepgfplotslibrary{groupplots,dateplot}
\usetikzlibrary{patterns,shapes.arrows}
\pgfplotsset{compat=newest}

\newcommand{\ours}{{RARG}\xspace}

\title{Evidence-Driven Retrieval Augmented Response Generation for Online Misinformation}

\author{Zhenrui Yue \quad Huimin Zeng \quad Yimeng Lu \quad Lanyu Shang \\
\textbf{Yang Zhang \quad Dong Wang} \\
University of Illinois Urbana-Champaign \\
\texttt{\{zhenrui3, huiminz3, yimengl4, lshang3, yzhangnd, dwang24\}@illinois.edu} \\}

\begin{document}
\maketitle
\begin{abstract}
The proliferation of online misinformation has posed significant threats to public interest. While numerous online users actively participate in the combat against misinformation, many of such responses can be characterized by the lack of politeness and supporting facts. As a solution, text generation approaches are proposed to automatically produce counter-misinformation responses. Nevertheless, existing methods are often trained end-to-end without leveraging external knowledge, resulting in subpar text quality and excessively repetitive responses. In this paper, we propose retrieval augmented response generation for online misinformation (\ours), which collects supporting evidence from scientific sources and generates counter-misinformation responses based on the evidences. In particular, our \ours consists of two stages: (1)~evidence collection, where we design a retrieval pipeline to retrieve and rerank evidence documents using a database comprising over 1M academic articles; (2)~response generation, in which we align large language models (LLMs) to generate evidence-based responses via reinforcement learning from human feedback (RLHF). We propose a reward function to maximize the utilization of the retrieved evidence while maintaining the quality of the generated text, which yields polite and factual responses that clearly refutes misinformation. To demonstrate the effectiveness of our method, we study the case of COVID-19 and perform extensive experiments with both in- and cross-domain datasets, where \ours consistently outperforms baselines by generating high-quality counter-misinformation responses. 
\end{abstract}

\input{1_intro}
\input{2_related}
\input{3_method}
\input{4_exp}
\input{5_con}

\bibliography{anthology,custom}

\onecolumn
\appendix
\input{6_app}

\end{document}

%% file: 1_intro.tex
\begin{figure}[ht]
    \centering
    \includegraphics[trim=0 3.15cm 0 0.85cm, clip, width=1.0\linewidth]{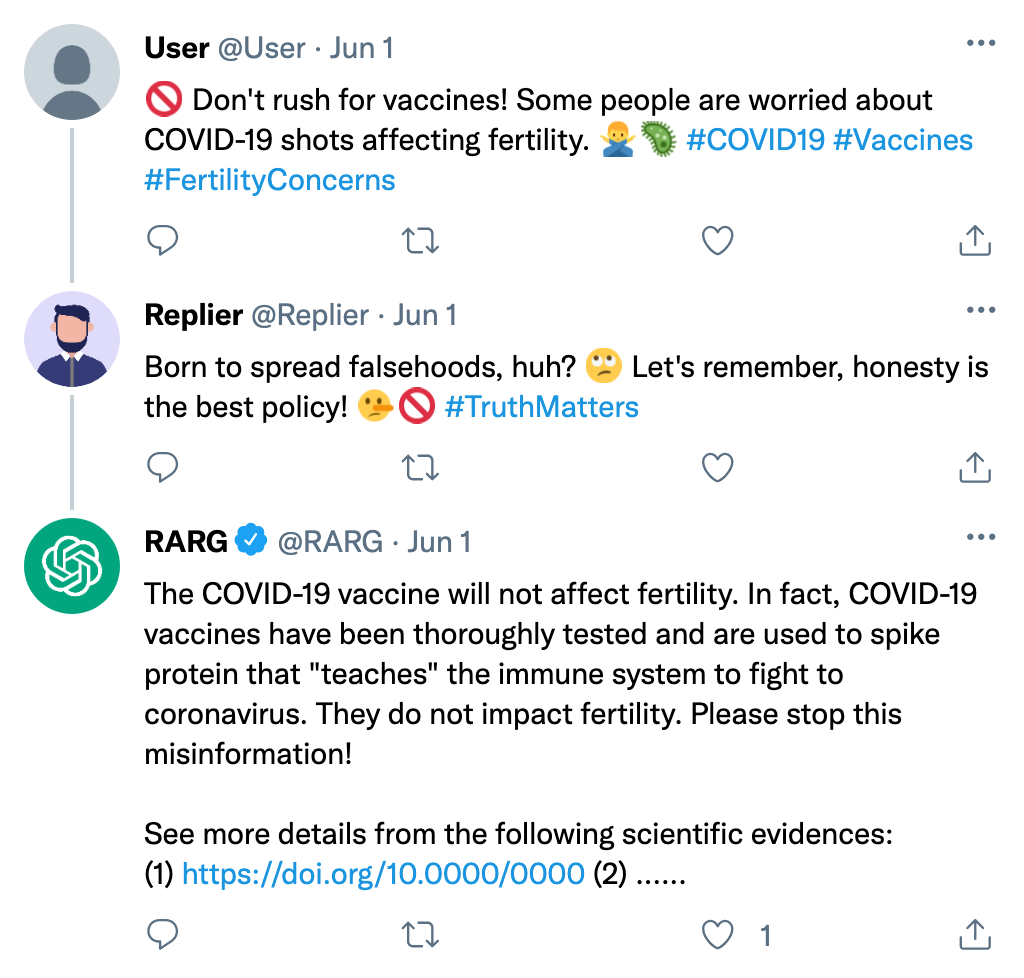}
    \caption{The proposed retrieval augmented response generation (\ours), which collects scientific evidence to generate counter-misinformation responses.} 
    \label{fig:intro}
\end{figure}

\section{Introduction}

As social media discussions on trending topics continue to grow, increased presence of online misinformation has been observed on such sources~\cite{chen2022combating, chen2023combating}. Yet timely intervention against the spreading of misinformation is often lacking, resulting in potential threats to public interest~\cite{ball2020epic, roozenbeek2020susceptibility}. For instance, vaccine hesitancy is strongly associated with the exposure of anti-vaccine misinformation~\cite{pierri2022online}. Therefore, it is essential to identify and curb the spreading of misinformation before it leads to severe consequences~\cite{litou2017efficient, zhu2021robust}. To identify text-based misinformation, various approaches have been proposed by leveraging language models~\cite{shu2020disinformation, wu-etal-2022-cross, shu2022cross, yue2022contrastive, yue-etal-2023-metaadapt, chen2023can, liu2024teller}. Upon the detection of misinformation, crowdsourcing approaches (e.g, users, fact-checkers) are designed to prevent the continued propagation of misinformation via counter-responses~\cite{veeriah2021young, vraga2021addressing, kou2022hc, seo2022if, kou2022crowd, bozarth2023wisdom, drolsbach2023diffusion}. Nevertheless, such responses to misinformation often lack politeness, direct refutation or fail to provide supporting evidence~\cite{he2023reinforcement, tafur2023user}. Moreover, human efforts are often less effective when confronted with an overwhelming amount of online misinformation~\cite{micallef2020role, micallef2022cross}.

To facilitate early intervention at scale, language generation methods are proposed to generate counter-responses against detected misinformation~\cite{vo2019learning, vo2020standing, he2023reinforcement}. For example, MisinfoCorrect adopts reinforcement learning to generate responses to refute misinformation~\cite{he2023reinforcement}. Despite their effectiveness, existing generation methods are trained end-to-end on a specific domain. As a result, the generated texts often highly resemble the training responses and demonstrate deteriorated quality upon domain shifts (as we show in \Cref{sec:exp}). Additionally, such generation models are unaware of external knowledge and must be frequently updated to incorporate up-to-date domain knowledge~\cite{lewis2020retrieval, izacard-grave-2021-leveraging, borgeaud2022improving, asai2023self}. As such, we consider a retrieval augmented generation (RAG) setting that incorporates relevant scientific documents to utilize external evidence without retraining. That is, given input misinformation, our first objective is to retrieve evidence from an extensive collection of documents (e.g., academic publications). Then, we select the relevant documents as evidence to perform retrieval augmented generation, with the goal of extracting supporting facts to debunk misinformation. We illustrate an example of our framework in \Cref{fig:intro}, where evidence-backed counter-response is generated upon the identification of misinformation.

In this paper, we focus on counter-response generation against online misinformation and propose an evidence-driven retrieval augmented response generation (\ours) framework, which efficiently retrieves supporting documents and generates responses using the collected supporting facts. Specifically, \ours consists of two modules: (1)~evidence collection, in which we design a two-stage retrieval pipeline based on a collection of over 1M academic articles we collected. For efficient retrieval, our pipeline first performs a coarse search over the data collection, followed by a reranking stage with improved relevance estimation of the retrieved documents; (2)~evidence-based response generation, here, we align large language models (LLMs) to generate responses upon the collected evidence via reinforcement learning from human feedback (RLHF). In particular, we propose a reward design that maximizes the utilization of the retrieved evidence while maintaining the quality of the responses. As such, our \ours generates polite and factual responses that clearly refutes input misinformation. To the best of our knowledge, we are the first to introduce a retrieval augmented generation framework with two-stage retrieval and fine-grained RLHF. To demonstrate the effectiveness of \ours, we study the case of COVID-19, where we perform both in- and cross-domain experiments with comprehensive quantitative and qualitative analyses. Experimental results highlight the effectiveness of the proposed framework, where \ours consistently outperforms state-of-the-art baselines by generating high-quality evidence-based responses against online misinformation. 

We summarize our contributions as follows:
\begin{enumerate}
\item We propose a retrieval augmented generation setting for counter-misinformation response generation. In this work, we focus on COVID-19 misinformation and collected over 1M academic publications as the source for evidence-based response generation.

\item We design \ours, a response generation framework against online misinformation. Our framework combines two-stage retrieval for evidence collection and RLHF-based LLM alignment, such that \ours is optimized to generate polite and factual counter-responses.

\item We show the effectiveness of \ours by experimenting on both in-domain and cross-domain COVID misinformation to validate the generalization of \ours. Both quantitative and qualitative results demonstrate that \ours can outperform state-of-the-art methods in generating high-quality responses.
\end{enumerate}

%% file: 2_related.tex
\section{Related Work}

\subsection{Detecting and Countering Misinformation}
Existing methods for detecting misinformation can be broadly categorized into: (1)~content-based detection, in which machine learning models are trained upon input contents (i.e., claims) to perform classification~\cite{yue2022contrastive, zeng2022unsupervised, jiang2022fake, yue-etal-2023-metaadapt, chen2023can, liu-etal-2023-interpretable, mendes-etal-2023-human, zeng2024unsupervised, liu2024teller}. Additional modalities such as image, video or propagation paths can also be used to improve detection performance~\cite{shang2021multimodal, santhosh2022multi, shang2022duo, wu-etal-2022-cross, zhou2023multimodal}; (2)~evidence-based detection, where external knowledge can be collected as supporting evidence to verify input contents~\cite{brand2021bart, kou2022hc, wu2022bias, yang-etal-2022-coarse, shang2022knowledge, xu2022evidence, zhao-etal-2023-panacea}. For example, knowledge graphs or retrieved document pieces can be processed to support or refute statements~\cite{kou2021fakesens, hu-etal-2021-compare, koloski2022knowledge, kou2022crowd, shang2022privacy, wu2022adversarial}. 

To curb the spreading of misinformation, the majority of existing approaches relies on social correction (e.g., comments, replies) by users or experts~\cite{veeriah2021young, vraga2021addressing, seo2022if, bozarth2023wisdom}. Despite their effectiveness, many of such responses can be characterized by the lack of politeness and supporting facts. Furthermore, user efforts often fall short when confronted with an overwhelming amount of misinformation~\cite{micallef2020role, micallef2022cross, he2023reinforcement}. Recently, language-based methods are proposed to generate counter-responses~/~explanations~\cite{vo2019learning, vo2020standing, he2023reinforcement, wan2024dell}. However, these methods concentrate on improving text quality, often neglecting the response factuality through the incorporation of evidence. Hence, we aim to design a counter-misinformation response generation framework via the collection of relevant documents, followed by the reasoning and generation of responses.

\subsection{Retrieval Augmented Generation}
Recent progress in large language models (LLMs) has led to substantial improvements in both language understanding and generation~\cite{raffel2020exploring, brown2020language, wei2021finetuned, ouyang2022training, chowdhery2022palm, touvron2023llama, openai2023gpt}. Thanks to the extensive corpora used during pretraining, LLMs have the ability to embed world knowledge in their parameters, and thus achieve significant performance enhancements across various scenarios~\cite{touvron2023llama, openai2023gpt, penedo2023refinedweb}. Nevertheless, LLMs struggle to capture fine-grained knowledge and frequently exhibit instances of hallucination~\cite{sun2023head, peng2023check}. To incorporate up-to-date knowledge without expensive retraining, retrieval augmented generation is proposed to generate text conditioned on collected documents~\cite{lewis2020retrieval, izacard-grave-2021-leveraging, borgeaud2022improving, izacard2022few, shi2023replug, ram2023context}. For example, REALM retrieves from a large set of documents such as Wikipedia to solve conditional generation tasks like question answering~\cite{guu2020retrieval}. Nevertheless, current retrieval augmented methods primarily study knowledge-intensive NLP tasks and have not been well researched for response generation against online misinformation, let alone the combination of fine-grained retrieval with RLHF-based alignment. As such, our work explores retrieval augmented response generation by collecting scientific evidence and performing RLHF alignment to generate evidence-based counter-responses.

%% file: 3_method.tex
\section{Methodology}

\subsection{Preliminary}
We consider the following problem setup for counter-misinformation response generation: given input misleading claim $x$, our objective is to: (1)~collect a set of $m$ scientific evidence $\{ e_i \}_{i=1}^{m}$ that are relevant to input $x$ to be used as supporting evidence; and (2)~generate response $y$ upon input $x$ and the retrieved evidence $\{ e_i \}_{i=1}^{m}$, which should demonstrate certain desirable properties (e.g., politeness and factuality), see example in \Cref{fig:intro}. We elaborate our framework in the following.

\noindent
\textbf{Input \& Output}: We denote the evidence collection model as $f_{\mathrm{ret}}$ and the response generation model as $f_{\mathrm{gen}}$. Formally, our research framework can be defined as two sub-problems of information retrieval (i.e., evidence retrieval) and retrieval augmented text generation (i.e., response generation), with each of the settings defined as follows:
\begin{itemize}[leftmargin=10pt]
    \item \emph{Evidence Retrieval}: Given input misinformation $x$, human annotated evidence $e$ and a collection of $n$ evidence documents $\{ e_i \}_{i=1}^{n}$ (with $e \in \{ e_i \}_{i=1}^{n}$), the model $f_{\mathrm{ret}}$ should ideally generate the highest relevance score for the claim-evidence pair (i.e., $f_{\mathrm{ret}}(x, e) = \max \{ f_{\mathrm{ret}}(x, e_i) \}_{i=1}^{n}$). During training, input $x$ and $e$ can be used to optimize $f_{\mathrm{ret}}$. In inference, we collect a subset of $m$ documents $\{ e_i \}_{i=1}^{m}$ for response generation, with $m \ll n$.    
    \item \emph{Response Generation}: For response generation purpose, we incorporate both input claim $x$ and collected evidence subset $\{ e_i \}_{i=1}^{m}$ from the previous step in the input prompt, while $y$ represents the generated response from $f_{\mathrm{gen}}$ (i.e., $y = f_{\mathrm{gen}}(x, \{ e_i \}_{i=1}^{m})$). To learn the generation model, we first train $f_{\mathrm{gen}}$ end-to-end, followed RLHF tuning. The inference stage is completed by collecting evidence documents and performing evidence-based response generation.
\end{itemize}

\noindent
\textbf{Optimization:} The retrieval model $f_{\mathrm{ret}}$ and generation model $f_{\mathrm{gen}}$ are parameterized by $\theta_{\mathrm{ret}}$ and $\theta_{\mathrm{gen}}$. To learn $f_{\mathrm{ret}}$, we maximize the relevance score of the label evidence $e$ upon input $x$. In other words, we minimize the expected loss of input-evidence pair ($x, e$): {\small $\min_{\substack{\theta_{\mathrm{ret}}}} \mathbb{E}_{(x, e) \sim \mathcal{X}_{\mathrm{ret}}} [\mathcal{L}(\theta_{\mathrm{ret}}, (x, e))]$}, with $\mathcal{X}_{\mathrm{ret}}$ representing the evidence retrieval dataset. For $f_{\mathrm{gen}}$, the optimization is two-fold: (1)~$\theta_{\mathrm{gen}}$ is first trained end-to-end by minimizing cross entropy loss on input-evidence-response triplets, where $m$ can be empirically selected (i.e., {\small $\min_{\substack{\theta_{\mathrm{gen}}}} \mathbb{E}_{(x, \{e_i\}_{i=1}^{m}, y) \sim \mathcal{X}_{\mathrm{gen}}} [\mathcal{L}(\theta_{\mathrm{gen}}, (x, \{e_i\}_{i=1}^{m}, y))]$}); (2)~to improve the response quality, we leverage an additional reinforcement learning step to tune $\theta_{\mathrm{gen}}$ towards generating responses of human preference. 
In particular, we design the reward by considering the following aspects that could help reducing the spread of misinformation~\cite{starbird2014rumors, chan2017debunking, tanaka2019exposure, malhotra2022meaning, he2023reinforcement}:
\begin{itemize}[leftmargin=10pt]
    \item \emph{Refutation}: Refutation involves phrases within the response that explicitly and objectively expose the error or inaccuracy of input claims. 
    \item \emph{Factuality}: Factuality evaluates the correctness and supporting evidence of the response, which reflects the reliability of the generated text. 
    \item \emph{Politeness}: Politeness refers to using considerate and thoughtful language, and thus respectfully counters misinformation and avoids backfire.
    \item \emph{Claim Relevance}: Response should establish a clear and direct connection to the input claim for improved coherence and comprehension.
    \item \emph{Evidence Relevance}: The relevant evidence information should be included in the response to demonstrate the falsehood of input claim.
\end{itemize}
To summarize, we introduce a retrieval augmented response generation framework with two-stage evidence retrieval and fine-grained reinforcement learning, which learns to construct evidence-based counter-responses against misinformation. 

\begin{figure*}[t]
    \centering
    \includegraphics[trim=1.5cm 4.2cm 1.3cm 4.2cm, clip, width=1.0\linewidth]{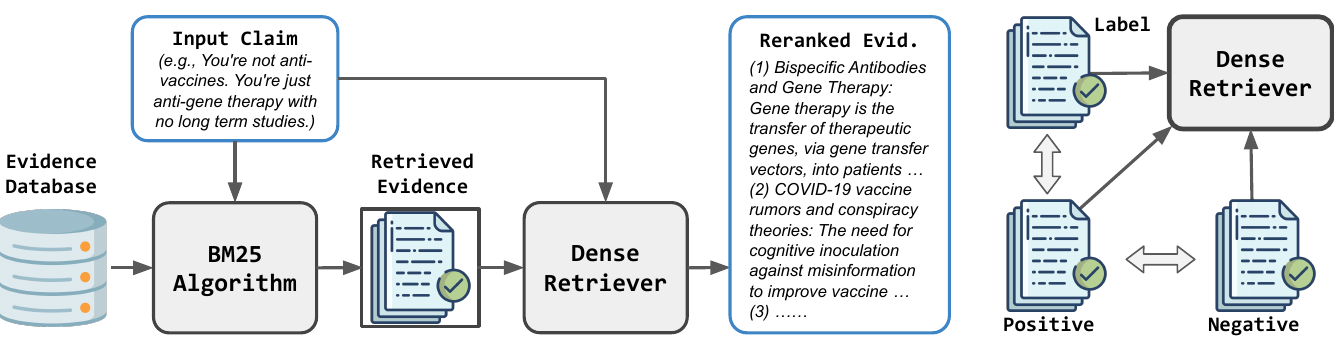}
    \caption{The proposed two-stage retrieval pipeline in \ours (left) and its optimization (right).}
    \label{fig:retrieval}
\end{figure*}

\subsection{Two-Stage Evidence Retrieval}
\label{sec:retrieval}
Existing retrieval augmented generation approaches adopts unsupervised sparse retrieval to collect relevant documents~\cite{lewis2020retrieval, izacard-grave-2021-leveraging, izacard2022few, ram2023context}. An example of such algorithms include BM25, which computes query-document relevance by considering word frequency in both query and documents~\cite{robertson1995okapi}. Although BM25 is widely used, it does not yield satisfactory ranking performance for knowledge-intensive tasks such as question answering. Therefore, dense retrieval methods are proposed for improved text understanding and relevance estimation~\cite{karpukhin-etal-2020-dense, ren-etal-2021-pair, izacard2021unsupervised, wang2022text, wang2023improving}. Nevertheless, dense retrieval is known to be inefficient for massive data quantity and require large amounts of annotated data for training. This limitation renders dense retrieval to be less effective in retrieval augmented response generation, where only limited annotated data is available for training.

Unlike existing retrieval methods, we design a two-stage retrieval pipeline in \ours that performs coarse-to-fine ranking, which improves the computation efficiency and retrieval performance with limited data. Specifically, we include the following stages: (1)~retrieve a smaller subset from the large collection of evidence documents via BM25; and (2)~rerank the retrieved subset using a dense retriever fine-tuned on limited claim-evidence pairs. In the first retrieval stage, we adopt BM25 to efficiently generate a small subset $\{ e_i \}_{i=1}^{m}$ from a much large collection $\{ e_i \}_{i=1}^{n}$. While BM25 may occasionally retrieve less or ir-relevant documents, we observe that in most cases, the desired evidence can still be found in $\{ e_i \}_{i=1}^{m}$ with proper selection of $m$. In our implementation, we adopt $m = 20$ as the subset size to balance the performance and efficiency of our retrieval pipeline. In the following stage, we fine-tune a dense retrieval model to perform fine-grained reranking and select the most relevant documents as supporting evidence. Compared to sparse retrieval, this additional stage yields notable performance improvements with minimal computational overhead. We illustrate the overall retrieval pipeline of \ours in \Cref{fig:retrieval} (left).

For dense retriever $f_{\mathrm{den}}$ within $f_{\mathrm{ret}}$, fine-tuning often requires large amounts of annotated data for improved ranking performance and generalizability~\cite{karpukhin-etal-2020-dense, wang2022text}. Despite the insufficiency of data, the retrieval results from our first stage can be used as a coarse estimation of claim-evidence relevance. In other words, it is possible to leverage the BM25 scores from the first stage, combined with limited annotated examples, to optimize the dense retrieval model. Specifically for input-evidence pair $(x, e)$, we sample $k$ positive documents $\{ e_i^p \}_{i=1}^{k}$ and $k$ negative documents $\{ e_i^n \}_{i=1}^{k}$ using BM25 and inverse BM25 scores, which avoids introducing extensive noise in the training data. Based on the positive set $\{ e_i^p \}_{i=1}^{k}$, we design a ranking loss to enlarge the margin between evidence $e$ and the highest ranked positive evidence in $\{ e_i^p \}_{i=1}^{k}$ (i.e., {\small $f_{\mathrm{den}}(x, e) - \max \{ f_{\mathrm{den}}(x, e_i^p) \}_{i=1}^{k})$}). Here, we maximize the relevance between input-evidence pairs by applying penalty only when the margin is below threshold $\tau$. In addition to the ranking loss, our optimization goal contains a contrastive term based on InfoNCE loss~\cite{chen2020simple}, which improves the relevance estimation between input-evidence pairs while `pushing away' negative evidence. Formally, the overall optimization objective $\mathcal{L}$ for $f_{\mathrm{den}}$ can be formulated as:
\begin{equation}
    \small
    \begin{aligned}
        \mathbb{E}_{(x, e) \sim \mathcal{X}_{\mathrm{ret}}} [ & \max (0, \max (\{ f_{\mathrm{den}}(x, e_i^p) \}_{i=1}^{k}) - f_{\mathrm{den}}(x, e) + \tau) \\
        & - \lambda \frac{\exp(f_{\mathrm{den}}(x, e))}{\exp(f_{\mathrm{den}}(x, e)) + \sum_{k} \exp(f_{\mathrm{den}}(x, e_i^n))} ],
    \label{eq:retriever-loss}
    \end{aligned}
\end{equation}
where $\tau$ is the margin threshold for the ranking loss and $\lambda$ is a scaling factor. For each pair of $x$ and $e$, the first term in \Cref{eq:retriever-loss} is effective when $f_{\mathrm{den}}(x, e)$ score is not $\tau$ greater than the score of the highest ranked positive evidence. The next term, on the other hand, maximizes exponential score of the annotated input-evidence pair over the sum of those from the sampled negative evidence. We illustrate the optimization of the dense retriever in \Cref{fig:retrieval} (right), where we differentiate the annotated evidence $e$ from the positive evidence $\{ e_i^p \}_{i=1}^{k}$ and `push away' the negative evidence $\{ e_i^n \}_{i=1}^{k}$.

For the evidence document collection, since we focus on the case of COVID-19, we limit the evidence documents to academic articles on COVID research from reliable sources. In particular, we collect all available articles from two large databases of COVID-related research: CORD~\cite{wang2020cord} and LitCovid~\cite{ chen2021litcovid}. In sum, we obtain 1,056,262 articles from CORD and 301,136 articles from LitCovid\footnote{CORD is maintained until June 2022, while LitCovid is active as of October 2023. Our LitCovid collection is updated with information available until August 2023.}. The collected articles are preprocessed to filter invalid and repeated items. As a result, 1,118,112 articles remain in our evidence database and we extract title and abstract information from each article as the evidence corpus. To fine-tune our dense retriever model, we utilize Check-COVID, a dataset with $\sim$350 evidence documents and $\sim$1k input-evidence pairs for training, with all documents collected from the CORD dataset~\cite{wang-etal-2023-check-covid}. To improve the ranking performance over large evidence collections, we manually increase the document size of Check-COVID to 5k by sampling additional documents from CORD and LitCovid. We use $k = 4$ in $\{ e_i^p \}_{i=1}^{k}$ and $\{ e_i^n \}_{i=1}^{k}$, we also provide more data collection and processing details in \Cref{sec:exp} and \Cref{sec:exp_app}.

\begin{figure*}[t]
    \centering
    \includegraphics[trim=1.6cm 3cm 1.6cm 3.2cm, clip, width=1.0\linewidth]{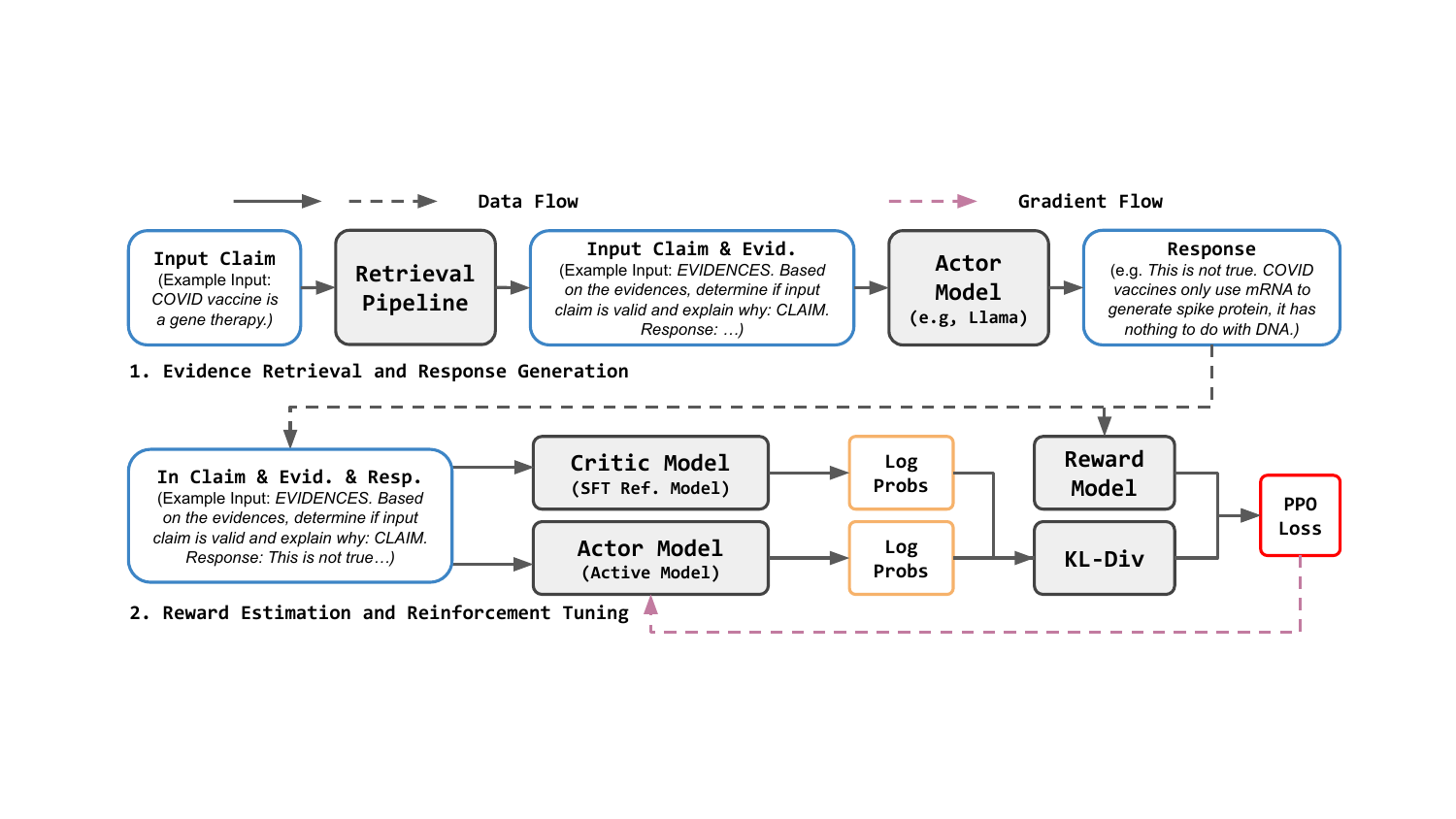}
    \caption{The optimization of $f_{\mathrm{gen}}$ in \ours. Upon input claim, the evidence documents are retrieved for response generation. Then, the reward model estimates the rewards and update the actor model with PPO-based RL.}
    \label{fig:generation}
\end{figure*}

\subsection{Response Generation}
\label{sec:generation}
Despite recent progress in large language models (LLMs), it is frequently observed that LLMs demonstrate hallucination behavior due to the lack of knowledge and reasoning capabilities~\cite{sun2023head, peng2023check}. To improve the factuality of LLM-generated text, retrieval augmented text generation is used to incorperate external knowledge~\cite{lewis2020retrieval, izacard-grave-2021-leveraging, borgeaud2022improving, ram2023context}. Yet the training of such models heavily relies on the quality of the data and demands extensive tuning efforts to achieve substantial performance improvements. On the other hand, generating large amounts of responses with desired properties (e.g., factuality, politeness) can be time-consuming and costly. In contrast, gathering basic human feedback such as preference ranking proves to be a notably simpler and more cost-effective approach for feedback collection~\cite{ouyang2022training}. Therefore, we combine the advantages of LLMs, which can be easily instruction-tuned with small-size data, while exploiting the potential for further improvement through RLHF~\cite{ouyang2022training}.

To optimize the generation model, we propose a RLHF-based approach to generate responses towards human preference. Specifically, $f_{\mathrm{gen}}$ is first fine-tuned in a supervised fashion to generate counter-response conditioned on input claim and retrieved evidence. At the same time, a reward model is constructed by leveraging binary human feedback (e.g., whether responses are refuting, factual and polite) as training data to evaluate the response quality. The reward model is then applied in the reinforcement learning stage, where the fine-tuned $f_{\mathrm{gen}}$ (i.e., actor model) is trained to further improve generation quality using proximal policy optimization (PPO)~\cite{schulman2017proximal}. Upon deployment, only the actor model is required for inference, which generates counter-misinformation responses based on input claims and collected evidence. We provide an illustration for the response generation process and its optimization in \Cref{fig:generation}.

\textbf{Supervised Fine-Tuning.} Reinforcement learning is known for being unstable for complex tasks, therefore, we first perform supervised fine-tuning using a pretrained LLM (e.g., Llama 2) to learn initial capabilities in counter-misinformation response generation. Additionally, the fine-tuned model is used as a reference model (i.e., critic model in \Cref{fig:generation}) in the following reinforcement learning step to stabilize training. We denote the supervised fine-tuned model as $\pi_{\mathrm{ref}}$ (i.e. reference model).

\textbf{Modeling Reward.} The purpose of reward modeling is to leverage human feedback for evaluating the text quality. Unlike methods that generate text pairs for preference evaluation~\cite{ouyang2022training}, we train smaller LMs to model human feedback, as the swift convergence accelerates reward estimation in RLHF. For a dataset of $\{ x_i, y_i, r_i \}_{i=1}^{n}$ with $r \in \{ 0, 1 \}$ representing the reward label from human feedback (i.e., refutation, factuality or politeness in binary form), the objective of modeling reward is to learn a modeling function $f$ that minimizes the expected negative log-likelihood loss. The learnt reward model evaluates the quality of the generated responses. In our implementation, we adopt three BERT models to learn human feedback in refutation, factuality, politeness, and use the sum of the scores as reward. To further improve the coherence of the response w.r.t. input claim and retrieved evidence, we assess the relevance between claim-response pairs (i.e, $(x, y)$) and evidence-response pairs (i.e, $(e, y)$). To maximize parameter efficiency, we utilize $f_{\mathrm{den}}$ from the previous retrieval module, since $f_{\mathrm{den}}$ is specifically trained to model the evidence relevance. Thus the estimated reward $\hat{r}$ is formulated as:
\begin{equation}
    \small
    \begin{aligned}
        \hat{r} & = f_{\mathrm{refutation}}(x, \{ e_i \}_{i=1}^{m}, y) + f_{\mathrm{factuality}}(x, \{ e_i \}_{i=1}^{m}, y) \\
        & + f_{\mathrm{politeness}}(x, \{ e_i \}_{i=1}^{m}, y) + \alpha (f_{\mathrm{den}}(x, y) \\
        &  + \max \{ f_{\mathrm{den}}(e_i, y) \}_{i=1}^{m}),
    \label{eq:reward-construction}
    \end{aligned}
\end{equation}
where $f_{\mathrm{refutation}}$, $f_{\mathrm{factuality}}$ and $f_{\mathrm{politeness}}$ model refutation, factuality and politeness respectively. $f_{\mathrm{den}}(x, y)$ is used to evaluate claim-response relevance, and $\max \{ f_{\mathrm{den}}(e_i, y) \}_{i=1}^{m}$ evaluates the best-matching evidence-response relevance. $\alpha$ is a hyperparameter to scale the relevance reward.

\textbf{Reinforcement Learning.} For reinforcement learning with human feedback, we use \Cref{eq:reward-construction} to assess the quality of the generated responses. The objective is to maximize the expected reward w.r.t. the actor model under the constraint that the parameters in $\pi_{\mathrm{act}}$ do not significantly deviate from $\pi_{\mathrm{ref}}$. Formally, the learning is formulated as:
\begin{equation}
    \small
    \begin{aligned}
        \max_{\pi_{\mathrm{act}}} & \mathbb{E}_{(x, e) \sim \{ x_i, e_i \}_{i=1}^{n}, y \sim \pi_{\mathrm{act}} (x, e)} [ \hat{r}(x, e, y) ] \\
        & - \beta \mathbb{D}  [ \log \pi_{\mathrm{act}} (y | x, e) || \log \pi_{\mathrm{ref}} (y | x, e ) ]],
    \label{eq:rl-formulation}
    \end{aligned}
\end{equation}
where $\beta$ is a hyperparameter to regularize the output difference between $\pi_{\mathrm{act}}$ and $\pi_{\mathrm{ref}}$, and $\mathbb{D}$ stands for the KL divergence. Here, we aim to find the optimal policy $\pi_{\mathrm{act}}$ that maximizes the expected reward $\hat{r}$. The additional KL divergence term controls how far the actor model can travel from the reference model via the minimization of KL divergence. Therefore, penalizing the KL distance effectively prevents optimization instability or model collapse. During training, we initialize the actor model $\pi_{\mathrm{act}}$ with reward head using the weights from $\pi_{\mathrm{ref}}$ and leverage PPO as the learning algorithm~\cite{schulman2017proximal, ouyang2022training}. We provide an illustration of the training pipeline in \Cref{fig:generation}, where the upper subfigure demonstrates evidence retrieval and response generation. The lower subfigure explains our reinforcement learning scheme, in which the generated responses are used to estimate the rewards and compute the KL distances. Finally, the rewards and KL penalty are used the compute the training loss and update the actor model. After the reinforcement learning stage, the resulting actor model $\pi_{\mathrm{act}}$ is used as our response generation model in \ours.

%% file: 4_exp.tex
\section{Experiments}
\label{sec:exp}

\subsection{Experiment Design}
\textbf{Evidence Retrieval.} Our retrieval module consists of BM25 and $f_{\mathrm{den}}$ (initialized with E5~\cite{wang2022text}). We adopt the Check-COVID dataset for evidence retrieval evaluation~\cite{wang-etal-2023-check-covid}, the adopted metrics are NDCG and Recall (i.e., N@$k$ and R@$k$) with $k \in [1, 3, 5]$. For baselines, we adopt sparse algorithms TFIDF and BM25~\cite{robertson1995okapi}. We also incorporate state-of-the-art dense retrievers DPR and E5 for comparison~\cite{karpukhin-etal-2020-dense, wang2022text}.

\noindent
\textbf{Response Generation.} We adopt our collected scientific articles (see \Cref{sec:retrieval}) and use the top-3~/~top-5 documents from retrieval. Our $f_{\mathrm{gen}}$ is based on Llama 2 (7B)~\cite{touvron2023llama}, with the supervised fine-tuning variant denoted by \ours (S). We adopt MisinfoCorrect dataset for training~\cite{he2023reinforcement} and perform both in-domain and cross-domain evaluation (on Constraint and ANTiVax)~\cite{patwa2021fighting, hayawi2022anti}. We follow the evaluation from~\cite{he2023reinforcement} using metrics refutation (R.), factuality (F.) and politeness (P.), we also evaluate the response relevance to input claim (C.) and evidence (E.) using our dense retriever (i.e., $f_{\mathrm{den}}$). Baseline methods include BART, DialoGPT, PARTNER, GODEL, MisinfoCorrect (MisinfoC.), Llama 2 w/ and w/o retrieval (denoted with Llama and Llama (R)) and GPT-3.5~\cite{lewis-etal-2020-bart, zhang-etal-2020-dialogpt, sharmatowards, peng2022godel, he2023reinforcement, touvron2023llama, brown2020language}.

\begin{table}[t]
\footnotesize
\centering
\begin{tabularx}{1.0\linewidth}{Xccccc}
\toprule
\multirow{2}{*}{} & \multicolumn{5}{c}{\textbf{Check-COVID Dataset}} \\ \cmidrule(l){2-6} 
                  & N@1 $\uparrow$ & N@3 $\uparrow$ & R@3 $\uparrow$ & N@5 $\uparrow$ & R@5 $\uparrow$ \\ \midrule
\textbf{TFIDF}    & 0.266          & 0.363          & 0.427          & 0.385          & 0.480          \\
\textbf{BM25}     & 0.292          & 0.395          & 0.467          & 0.426          & 0.545          \\
\textbf{DPR}      & 0.324          & 0.411          & 0.477          & 0.457          & 0.588          \\
\textbf{E5}       & \ul{0.445}     & \ul{0.584}     & \ul{0.679}     & \ul{0.609}     & \ul{0.741}     \\ \midrule
\textbf{\ours}    & \textbf{0.513} & \textbf{0.631} & \textbf{0.712} & \textbf{0.646} & \textbf{0.750} \\ 
\bottomrule
\end{tabularx}
\caption{Evidence retrieval results, with best results in bold and second best results underlined.}
\label{tab:retrieval}
\end{table}

\begin{figure}[t]
    \centering
    \includegraphics[width=1.0\linewidth]{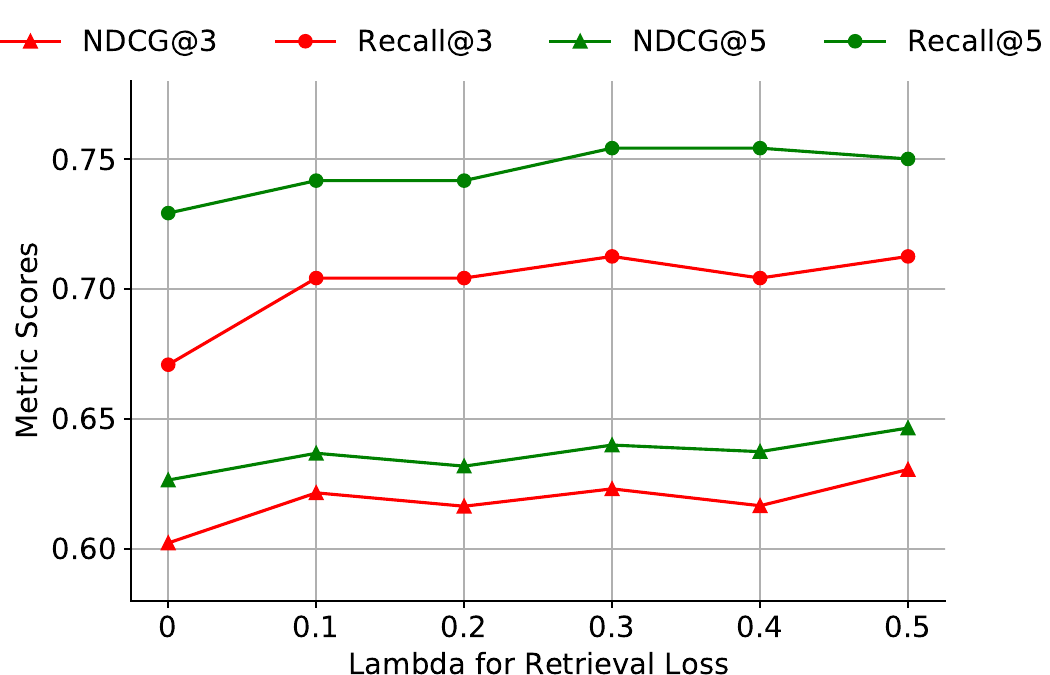}
    \caption{Hyperparameter sensitivity of $\lambda$ in retrieval.}
    \label{fig:lambda-sensitivity}
\end{figure}

\subsection{Evidence Retrieval Results}
Our main retrieval results are reported in \Cref{tab:retrieval}. In this table, rows represent retrieval methods and the columns represent different metrics. For top-1 scores, we use N@1 since top-1 NDCG and Recall scores are equivalent. From the results we observe: (1)~\ours retriever consistently outperforms baseline retrieval methods across all metrics, with an average performance improvement of $7.09\%$ compared to the second best results. (2)~In contrast to sparse retriever along, the additional dense retriever significantly improves the ranking performance. For example, RARG achieves $37.61\%$ performance improvement in Recall@5 compared to BM25. (3)~The performance gains through our dense retriever increases as we narrow the size of the retried subset (i.e., top-$k$). For instance, the NDCG improvements compared to the best baseline rise from $6.07\%$ to $8.04\%$ with $k$ decreasing from 5 to 3. (4)~By leveraging the ranking margin and contrastive learning, our dense retriever successfully exploits noisy BM25 results and outperforms the second best baseline E5 by $15.28\%$ in top-1 score, indicating strong reranking performance with \ours. Overall, we find the two-stage retrieval pipeline in \ours performs well even with limited training data, which demonstrates substantially improved retrieval performance.

\begin{table}[t]
\footnotesize
\centering
\begin{tabularx}{\linewidth}{Xccccc}
\toprule
\multirow{2}{*}{}      & \multicolumn{5}{c}{\textbf{MisinfoCorrect Dataset}} \\ \cmidrule{2-6}
                       & R. $\uparrow$       & F. $\uparrow$        & P. $\uparrow$        & C. $\uparrow$        & E. $\uparrow$    \\ \midrule
\textbf{BART}          & 0.824               & 0.623                & 0.824                & 0.799                & 0.685            \\
\textbf{DialoGPT}      & 0.831               & 0.693                & 0.874                & 0.800                & 0.689            \\
\textbf{PARTNER}       & 0.792               & 0.779                & 0.790                & 0.759                & 0.663            \\
\textbf{GODEL}         & \ul{0.931}          & 0.904                & 0.987                & 0.751                & 0.680            \\
\textbf{MisinfoC.}     & 0.916               & 0.914                & 0.927                & 0.788                & 0.686            \\
\textbf{Llama}         & 0.727               & 0.950                & 0.984                & 0.761                & 0.674            \\
\textbf{Llama} (R)     & 0.761               & 0.942                & 0.984                & 0.776                & 0.694            \\ 
\textbf{GPT-3.5}       & 0.857               & \textbf{0.971}       & 0.987                & \ul{0.805}           & 0.690            \\ \midrule
\textbf{\ours} (S)     & 0.922               & 0.944                & \ul{0.988}           & 0.804                & \ul{0.702}       \\
\textbf{\ours}         & \textbf{0.965}      & \ul{0.967}           & \textbf{0.989}       & \textbf{0.812}       & \textbf{0.704}   \\ \bottomrule
\end{tabularx}
\caption{In-domain response generation results, with best results in bold and second best results underlined.}
\label{tab:generation-indomain}
\end{table}

\begin{table*}[t]
\footnotesize
\centering
\begin{tabularx}{\linewidth}{lXXXXXcXXXXX}
\toprule
                        & \multicolumn{5}{c}{\textbf{Constraint Dataset}}                                            &|& \multicolumn{5}{c}{\textbf{ANTiVax Dataset}}                                               \\ \cmidrule(l){2-12} 
                        & R. $\uparrow$   & F. $\uparrow$    & P. $\uparrow$   & C. $\uparrow$    & E. $\uparrow$    &|& R. $\uparrow$   & F. $\uparrow$    & P. $\uparrow$   & C. $\uparrow$    & E. $\uparrow$ \\ \midrule
\textbf{BART}           & 0.949           & \textbf{0.969}   & \textbf{0.987}  & 0.692            & 0.608            &|& 0.942           & 0.949            & 0.960           & 0.741            & 0.641            \\
\textbf{DialoGPT}       & \ul{0.953}      & 0.953            & 0.968           & 0.636            & 0.584            &|& 0.952           & 0.952            & 0.966           & 0.748            & 0.641            \\
\textbf{PARTNER}        & 0.952           & 0.953            & 0.968           & 0.633            & 0.585            &|& 0.887           & 0.882            & 0.901           & 0.732            & 0.632            \\
\textbf{GODEL}          & 0.938           & 0.908            & \ul{0.986}      & 0.594            & 0.578            &|& 0.931           & 0.906            & 0.985           & 0.706            & 0.668            \\
\textbf{MisinfoC.} & \textbf{0.955}  & \ul{0.955}       & 0.970           & 0.636            & 0.584            &|& 0.952           & 0.952            & 0.967           & 0.747            & 0.641            \\ 
\textbf{Llama}        & 0.595           & 0.929            & 0.977           & 0.745            & 0.628            &|& 0.655           & 0.942            & 0.984           & 0.753            & 0.663            \\ 
\textbf{Llama} (R)  & 0.708           & 0.930            & 0.976           & 0.750            & 0.634            &|& 0.722           & 0.943            & 0.984           & 0.775            & 0.689            \\ 
\textbf{GPT-3.5}        & 0.855           & 0.933            & 0.979           & 0.795            & \textbf{0.684}   &|& 0.909           & 0.948            & 0.987           & 0.780            & \ul{0.689}       \\ \midrule
\textbf{\ours} (S)    & 0.941           & 0.937            & 0.981           & \ul{0.799}       & 0.679            &|& \ul{0.966}      & \ul{0.975}       & \ul{0.989}      & \ul{0.791}       & 0.687            \\
\textbf{\ours}          & 0.951           & 0.940            & 0.982           & \textbf{0.805}   & \ul{0.682}       &|& \textbf{0.971}  & \textbf{0.978}   & \textbf{0.991}  & \textbf{0.795}   & \textbf{0.691}   \\ \bottomrule
\end{tabularx}
\caption{Cross-domain response generation results, with best results in bold and second best results underlined.}
\label{tab:generation-outdomain}
\end{table*}

We additionally study the effect of $\lambda$ in the training objective of the retriever model, which regularizes the strength of contrastive loss. In particular, we vary the value of $\lambda$ from 0 to 0.5 and evaluate the retrieval performance on NDCG and Recall with $k \in [3, 5]$. We focus on top-3 and top-5 scores as they are representative ranking scores and show consistent trends with changing $\lambda$. We present the performance visually in \Cref{fig:lambda-sensitivity}, with x-axis representing the $\lambda$ values and y-axis representing the ranking scores. We observe consistent improvements by applying the contrastive loss, followed by minor changes with futher increasing $\lambda$ values. The performance for all metrics remains robust even with the maximum value of 0.5. In sum, the proposed contrastive loss improves the performance of \ours retrieval, and performs quite robust regardless of hyperparameter selections.

\subsection{Response Generation Results}
For response generation, we leverage the retrieved documents as part of the input prompt to generate the counter-responses. We report the in-domain response generation results in \Cref{tab:generation-indomain}, with rows representing response generation models and columns representing different metrics. To evaluate the model generalization, we additionally perform response generation on cross-domain datasets Constraint and ANTiVax~\cite{patwa2021fighting, hayawi2022anti}, with results presented in \Cref{tab:generation-outdomain}. We also demonstrate the generation quality on cross-domain data via qualitative examples in \Cref{tab:qualitative-examples}.

From both in- and cross-domain evaluation results we observe: (1)~Both \ours versions perform well on in-domain data and achieve superior performance over baseline methods. For example, \ours outperforms the best baseline on the refutation metric with $3.65\%$ relative improvement. (2)~RLHF-tuned \ours can further improve generation quality, achieving an average $2.40\%$ performance improvement on refutation, factuality and politeness metrics compared to \ours (S). (3)~Despite significantly reduced size (i.e., 7B), \ours performs similarly or exceeds GPT-3.5 on all metrics. In particular, RARG significantly outperforms GPT-3.5 on refutation, which may be attributed to the misinterpretation of certain examples by GPT-3.5, as we demonstrate in \Cref{tab:qualitative-examples}. (4)~In cross-domain experiments (\Cref{tab:generation-outdomain}), although most models show comparable performance on refutation, factuality and politeness metrics, \ours can generate responses of substantially enhanced coherence and quality. In contrast, baseline methods exhibit signs of overfitting, leading to responses that highly resemble the training data (see \Cref{tab:qualitative-examples}). (5) Llama 2 and GPT-3.5 perform well in cross-domain response generation (\Cref{tab:generation-outdomain}). Yet they show significantly reduced refutation scores, likely stemming from the frequent misclassification (i.e., as if responding to valid information). (6)~Combined both quantitative and qualitative results on cross-domain data, \ours clearly outperforms baseline generation methods with significantly improved claim and evidence relevance as well as generation quality. Overall, we conclude that the RLHF-tuned \ours demonstrates enhanced claim understanding, evidence-based reasoning and response generation abilities. Furthermore, \ours exhibits superior performance in countering both in-domain and cross-domain misinformation, highlighting the potential of \ours in generating counter-misinformation responses across a wide range of real-world scenarios.

%% file: 5_con.tex
\section{Conclusion}
In this paper, we propose a novel evidence-driven retrieval augmented response generation framework \ours against online misinformation. To the best of our knowledge, \ours is the first to introduce evidence-backed response generation to counter misinformation. The proposed \ours comprises of: (1)~evidence retrieval, where evidence documents are efficiently collected and reranked; and (2)~response generation, in which \ours generates evidence-based counter-responses that are factual and polite. We demonstrate the effectiveness of \ours by performing extensive experiments on multiple datasets, where \ours can consistently generate evidence-based responses with improved quality over state-of-the-art baseline methods.

\section{Limitations}
Despite introducing \ours for evidence-based counter-response generation against online misinformation, we have not discussed the setting where test domains significantly differ from COVID (e.g., fake news), which may hinder the deployment of the proposed method for more generalized applications. In addition, we have not studied the case when the reliability of evidence sources can not be guaranteed (e.g., documents from unverifiable online sources), which may introduce inaccuracies and impact the validity of the generated responses. Due to the limited research scope and budgets, we have not conducted human evaluations to assess the overall response quality of \ours. Therefore, we plan to explore a more generalized and domain-adaptive solution with an improved evaluation protocol for counter-response generation in the future.

%% file: 6_app.tex
\section{Experiment Details}
\label{sec:exp_app}

\begin{table}[t]
\footnotesize
\centering
\begin{tabularx}{0.6\linewidth}{Xcccc}
\toprule
\textbf{Datasets}       & \textbf{\#Evid.} & \textbf{\#Claim} & \textbf{\#Resp.} & \textbf{\#Leng.} \\ \midrule
\textbf{Check-COVID}    & 5.6k             & 1,068/229        & N/A              & 19.4             \\
\textbf{MisinfoC.} & N/A              & 568/134          & 568              & 38.8             \\
\textbf{ANTiVax}        & N/A              & 500              & N/A              & 37.2             \\
\textbf{Constraint}     & N/A              & 500              & N/A              & 25.0             \\ \bottomrule
\end{tabularx}
\caption{Dataset statistics.}
\label{tab:dataset}
\end{table}

\subsection{Evidence Retrieval} 
Our two-stage retrieval consists of BM25 and a fine-tuned dense retriever. The dense retriever is initialized with the pretrained embeddings from bidirectional encoder representations (E5)~\cite{wang2022text}. We describe the experimental design below.
\begin{itemize}[leftmargin=10pt]
    \item \emph{Dataset}: For retrieval evaluation, we adopt the Check-COVID dataset that provides claim-evidence pairs~\cite{wang-etal-2023-check-covid}. The dataset statistics of the training/test set is provided in \Cref{tab:dataset}. 
    \item \emph{Baseline}: For baseline models, we adopt unsupervised sparse retrieval algorithms TFIDF and BM25~\cite{robertson1995okapi}. We also incorporate two state-of-the-art dense retrievers for comparison: dense passage retrieval (DPR) and E5 retriever~\cite{karpukhin-etal-2020-dense, wang2022text}.
    \item \emph{Evaluation}: In our evaluation, we adopt the following evaluation ranking metrics: normalized discounted cumulative gain (NDCG@$k$) and recall (Recall@$k$) with $k \in [1, 3, 5]$. We validate the model using the best NDCG@10 scores and rerank the retrieved results.
\end{itemize}
For the subsequent response generation experiments, we adopt our collected scientific articles as the evidence sources (see \Cref{sec:retrieval}) and collect the top-3~/~top-5 documents for each input claim.

\begin{table}[t]
\footnotesize
\centering
\begin{tabularx}{0.6\linewidth}{Xccccc}
\toprule
                       & \textbf{BA} $\uparrow$ & \textbf{Acc.} $\uparrow$ & \textbf{F1} $\uparrow$ & \textbf{Prec.} $\uparrow$ & \textbf{Rec.} $\uparrow$ \\ \midrule
\textbf{Refutation}    & 0.835                  & 0.800                    & 0.873                  & 0.978                     & 0.788                    \\
\textbf{Factuality}    & 0.928                  & 0.933                    & 0.942                  & 0.925                     & 0.961                    \\
\textbf{Politeness}    & 0.849                  & 0.881                    & 0.923                  & 0.941                     & 0.905                    \\ \bottomrule
\end{tabularx}
\caption{Reward modeling results.}
\label{tab:reward}
\end{table}

\subsection{Response Generation} 
For experiments in response generation, the base model is the 7B version of Llama 2~\cite{touvron2023llama}. To improve parameter efficiency, \ours is trained with 8bit quantization and LoRA~\cite{hu2021lora, dettmers2022llm}, we report the experiment design below.
\begin{itemize}[leftmargin=10pt]
    \item \emph{Dataset}: We adopt MisinfoCorrect for training~\cite{he2023reinforcement}. All models are trained with the identical training set and evaluated on a non-overlapping test set for unbiased assessment. To evaluate model generalization, we additionally adopt Constraint and ANTiVax, with 500 sampled examples each, see \Cref{tab:dataset}~\cite{patwa2021fighting, hayawi2022anti}. 
    \item \emph{Baseline}: For baseline models, we adopt multiple state-of-the-art baselines. In particular, we adopt text generation-based BART, DialoGPT, GODEL~\cite{lewis-etal-2020-bart, zhang-etal-2020-dialogpt, peng2022godel}, reinforcement learning-based methods PARTNER and MisinfoCorrect~\cite{sharmatowards, he2023reinforcement}, LLM-based Llama 2, Llama 2 with retrieval (with identical retrieval as in \ours, denoted with Llama (R)) and GPT-3.5~\cite{brown2020language}.
    \item \emph{Evaluation}: We follow the evaluation protocol from, with metrics in refutation, factuality and politeness using the supervised trained LMs in~\Cref{sec:generation}~\cite{he2023reinforcement}. We also evaluate the response relevance to input claims and the collected evidence using the relevance estimation model (i.e., dense retriever) as in~\Cref{sec:retrieval}. 
\end{itemize}
Besides \ours, we also report results of the supervised fine-tuned version of \ours (i.e., \ours (S)). For reward modeling, we train three BERT models on the binary human feedback data w.r.t refutation, factuality and politeness. As the label distribution is imbalanced, we adopt stratified sampling to split train/test sets and apply class-balanced cross entropy loss for training. Besides accuracy, F1, precision and recall, we also use balanced accuracy (BA) to evaluate class-balanced performance on human feedback classification. The evaluation results for reward modeling are in \Cref{tab:reward}. In summary, BERT models achieve well-balanced classification performance across all human feedback metrics, with average BA of 0.871 and accuracy of 0.871.

\subsection{Implementation}
For retrieval baseline methods, we follow the original works for implementation and hyperparameter selection. For \ours retrieval pipeline, we use the E5-base version for further tuning and train with AdamW optimizer for 5 epochs. We search the learning rate from [1e-5, 2e-5, 3e-5] and select both $\tau$ and $\lambda$ from $[0.1, 0.2, 0.3, 0.4, 0.5]$. Training is performed with 100 warm up steps and cosine learning rate scheduler. For sampling evidence, 4 positive evidence documents and 4 negative documents are adopt for training. We align the rest settings with the original E5 work~\cite{wang2022text}. For the generation pipeline, we adopt publicly available implementation from the original authors. To train our 7B version Llama 2, we adopt 8bit quantization and LoRA for parameter efficient tuning, which is less than $0.2\%$ of the original parameters in Llama 2~\cite{hu2021lora, dettmers2022llm}. We use consistent LoRa settings with 16 as LoRA dimension, 32 as LoRA $\alpha$ along with 0.05 for LoRA dropout. The LoRA target modules are the $Q$ and $V$ projection matrices. For supervised fine-tuning, we train 3 epochs with learning rate of 1e-4 and 0.03 percentage of the steps as warmup. For reinforcement tuning, we adopt PPO with 1e-5 learning rate and 0.2 as initial KL regularization. Training epochs is selected between 1 to 3, batch size is selected between 4 to 32, and we update parameters after 4 gradient accumulation steps. To construct the text input for instruction tuning, we prepend evidence as context, followed by an instruction that describes the task. We also incorporate the input misinformation in the prompt before the start of the response token. The resulting prompt template is:
\begin{displayquote}
\#\#\# \texttt{Instruction}

\texttt{\{evidence\}; Based on the above evidence, determine if the claim is valid and explain why: \{claim\}}

\#\#\# \texttt{Response}

\texttt{\{response\}}
\end{displayquote}
where \texttt{evidence} and \texttt{claim} are filled by the collected evidence and claim. For training, the annotated response of each data example follows. For inference, the \texttt{response} position is left empty for generation.

\section{Qualitative Results}

We provide additional qualitative examples to compare the response quality among different response generation methods, examples can be found in \Cref{tab:qualitative-examples}. Due to space limitations, we select GODEL, MisinfoCorrect, Llama 2 (R) and GPT-3.5 for comparison, as the generated responses from such models are of the highest quality. Since \ours (S) and \ours generate similar responses to most claims, we provide qualitative example with RLHF-tuned version of \ours. Overall, we notice high similarity in the responses from GODEL and MisinfoCorrect compared to the training data. For LLMs, we observe significantly increased response length by Llama 2 (R) and GPT-3.5, which may affect the non-redundancy and readability scores in the response quality evaluation. In addition, Llama 2 (R) and GPT-3.5 often demonstrate false responses (i.e., hallucination) by taking the input claim as valid. For example in the example 1, both Llama 2 and GPT-3.5 incorrectly identifies the claim as valid, and thus generates invalid responses to the original misinformation. Similarly in example 2, GPT-3.5 produces questionable rationales on the efficacy of masks in combating COVID, possibly due to the lack of claim understanding. Considering both the model size and efficiency, \ours shows improved response generation capabilities and can outperform baseline methods w.r.t. the response quality.

\begin{table*}[h]
\small
\centering
\begin{tabularx}{\linewidth}{Xl}
\toprule
\multicolumn{2}{c}{\textbf{Response Examples from Constraint}} \\ \midrule
\textbf{Claim}      & \emph{At a time when studies are linking air pollution with COVID-19 mortality rates for the most vulnerable in} \\
                    & \emph{society, the Trump Administration is rolling back environmental protections - allowing more poison into air.} \\ \midrule
\textbf{GODEL}      & The statement that the vaccine manipulates the DNA is incorrect. As verified by fact-checkers, the vaccine ... \\
\textbf{MisinfoC.}  & This is not correct. The vaccine is not related to COVID-19. The vaccine only uses mRNA for ... \\
\textbf{Llama} (R)  & The claim made in the statement is \textbf{valid}. There have been several studies conducted that suggest a ... \\
\textbf{GPT-3.5}    & The claim is \textbf{valid}. Studies have shown a correlation between air pollution and COVID-19 mortality rates ... \\
\textbf{\ours}      & This is not correct. The air pollution has nothing to do with the mortality rate. The mortality rate is related \\
                    & to the virus. The air pollution is not the cause of the virus. Please stop this misinformation. \\ \midrule 

\textbf{Claim}      & \emph{Says a warning label on a box of disposable masks shows that they are ineffective at protecting against} \\
                    & \emph{the spread of COVID-19.} \\ \midrule
\textbf{GODEL}      & The assertion is wrong. The vaccine is not the gene therapy and the nurse is not committing crime. The ... \\
\textbf{MisinfoC.}  & The vaccine is not a vaccine. It is safe, and it is well tested on many people. Please stop spreading ... \\
\textbf{Llama} (R)  & The claim that a warning label on a box of disposable masks shows that they are ineffective at protecting ... \\
\textbf{GPT-3.5}    & It is \textbf{impossible to determine the validity} of this claim without seeing the actual warning label on the ... \\
\textbf{\ours}      & The warning label is not correct. The masks are effective at protecting against the spread of COVID-19. \\ \midrule 

\multicolumn{2}{c}{\textbf{Response Examples from ANTiVax}} \\ \midrule

\textbf{Claim}      & \emph{People who don't care about Covid: You can't live in fear of the virus. Also people who don't care about} \\
                    & \emph{Covid: I'm not getting a vaccine. It's experimental. We don't know what's in it. Me: So about living in fear?} \\ \midrule
\textbf{GODEL}      & Your understanding about the experiment is inaccurate. The vaccine is not the gene therapy, and it does ... \\
\textbf{MisinfoC.}  & This is not correct. The vaccine is not experimental, and it is not the experimental gene therapy. It only ... \\
\textbf{Llama} (R)  & The claim that people who don't care about COVID-19 are living in fear of the virus is not valid. It is ... \\
\textbf{GPT-3.5}    & The claim is not valid. The statement highlights a contradiction in the attitudes of people who claim they ... \\
\textbf{\ours}      & This is not correct. The vaccine is not experimental. It is well tested and safe. \\ \midrule 

\textbf{Claim}      & \emph{Firstly, how long have you worked there? 2nd If you have been there more than 3 years you will have more} \\
                    & \emph{rights. 3rd If you don't want this untested vaccine, find examples of people who have had heart attacks and} \\
                    & \emph{died because of vaccine. 4th Mention your human rights} \\ \midrule
\textbf{GODEL}      & Your understanding is not correct. The vaccine is safe, and before we take it, it has been tested on many ... \\
\textbf{MisinfoC.}  & Your claim is not correct. The vaccine is not untested and has not been found to cause heart problems ... \\
\textbf{Llama} (R)  & The claim you provided is not valid and is based on a series of false assumptions and misleading statement ... \\
\textbf{GPT-3.5}    & This claim is not valid. It contains several unrelated statements and does not present a clear argument ... \\
\textbf{\ours}      & Please stop spreading misinformation. The vaccine is not untested. It has been tested on thousands of people. \\
                    & It is safe and effective. \\

\bottomrule
\end{tabularx}
\caption{Generated responses for baseline methods and \ours on Constraint and ANTiVax datasets.}
\label{tab:qualitative-examples}
\end{table*}